\documentclass{article}

\usepackage{float}
\usepackage{algorithm}

\usepackage{PRIMEarxiv}

\usepackage[utf8]{inputenc} % allow utf-8 input
\usepackage[T1]{fontenc}    % use 8-bit T1 fonts
\usepackage{hyperref}       % hyperlinks
\usepackage{url}            % simple URL typesetting
\usepackage{booktabs}       % professional-quality tables
\usepackage{amsfonts}       % blackboard math symbols
\usepackage{nicefrac}       % compact symbols for 1/2, etc.
\usepackage{microtype}      % microtypography
\usepackage{lipsum}
\usepackage{fancyhdr}       % header
\usepackage{graphicx}       % graphics

\usepackage{booktabs}
\usepackage{mhchem}
\usepackage{cite}

\usepackage{algpseudocode}
\usepackage{xcolor}
\usepackage{authblk}
\usepackage{mhchem}

\newcommand{\data}[1]{
\section*{Data availability}
\if@anonymous Removed for anonymity \else #1 \fi}

\graphicspath{{media/}}     % organize your images and other figures under media/ folder

\title{
Stochastic hierarchical data-driven optimization: application to plasma-surface kinetics
}

\author[1]{\textbf{José Afonso}}
\author[1]{\textbf{Vasco Guerra}}
\author[1]{\textbf{Pedro Viegas}\thanks{Corresponding author: pedro.a.viegas@tecnico.ulisboa.pt}\hspace{0.9ex}}

\affil[1]{Instituto de Plasmas e Fusão Nuclear, Instituto Superior Técnico, Av. Rovisco Pais 1, Lisbon, Portugal}

\begin{document}
\maketitle

\begin{abstract}
This work introduces a stochastic hierarchical optimization framework inspired by Sloppy Model theory for the efficient calibration of physical models. Central to this method is the use of a reduced Hessian approximation, which identifies and targets the stiff parameter subspace using minimal simulation queries. This strategy enables efficient navigation of highly anisotropic landscapes, avoiding the computational burden of exhaustive sampling. To ensure rigorous inference, we integrate this approach with a probabilistic formulation that derives a principled objective loss function directly from observed data. We validate the framework by applying it to the problem of plasma-surface interactions, where accurate modelling is strictly limited by uncertainties in surface reactivity parameters and the computational cost of kinetic simulations. Comparative analysis demonstrates that our method consistently outperforms baseline optimization techniques in sample efficiency. This approach offers a general and scalable tool for optimizing models of complex reaction systems, ranging from plasma chemistry to biochemical networks.
\end{abstract}

% keywords can be removed

\section{Introduction}

The predictive modeling of complex natural systems, ranging from biochemical networks \cite{Daniel2008} and chemical engineering to plasma physics, often relies on multiphysics simulations constrained by numerous free parameters. A prevalent feature of these systems is their \textit{sloppy} nature \cite{Transtrum2015, MAAS1992239}: while the model may possess high-dimensional parameter spaces, its relevant physical behavior typically evolves along a low-dimensional latent manifold \cite{Transtrum_2011}. This observation aligns with the \textit{manifold hypothesis} in Scientific Machine Learning \cite{cayton2005algorithms}, which posits that meaningful signal can be separated from unidentifiable degrees of freedom by identifying the system's intrinsic geometry. Consequently, the refinement of these scientific models constitutes a challenging, ill-conditioned inverse problem where determining physical parameters from limited data is hindered by the ruggedness of the optimization landscape and the computational cost of the simulator.

Optimization techniques for such problems generally follow one of two strategies regarding their use of statistical approximations. The first strategy aims to improve sample efficiency by constructing a global statistical surrogate (e.g., Gaussian Processes) of the objective function \cite{sutton2018, Jones1998, hansen2016}. This allows the optimization to proceed on a computationally cheap proxy rather than the expensive simulator. However, relying on a surrogate introduces the risk that the optimization process could be misguided towards a false minimum that exists on the proxy's response surface but not in the true physical landscape.

The alternative strategy is to query the simulator directly, treating it as the ground truth without attempting to fit a global statistical proxy \cite{sutton2018}. In this work, we opt for this direct approach, motivated by the manageable computational cost of many simulators \cite{Tejero-del-Caz2019}. In this intermediate regime, while exhaustive sampling remains prohibitive, the trade-off of accepting the approximation error of a statistical surrogate is often unjustified. Instead, the challenge shifts to prioritizing sample efficiency: finding the optimal fit with the minimum number of simulator queries. This focus is critical to ensure the framework's scalability to more complex models, where a single evaluation may be significantly more expensive. Additionally, a major hurdle to efficiency is that these simulators often provide neither analytical nor numerical gradients.

To achieve sample efficiency without surrogates, we operationalize the manifold hypothesis by exploring the synergies between physical modeling and representation learning \cite{Karniadakis2021, Transtrum2015}. Rather than performing a blind search, we frame the optimization as a geometric inference task: separating meaningful signals from unidentifiable degrees of freedom. This reasoning parallels the logic of data-driven autoencoders, where the objective is to learn a compressed representation that captures the dominant features of the system \cite{hinton2006reducing}. Crucially, explicitly targeting this latent structure addresses the severe low-data regime inherent to computationally expensive simulators; by distinguishing the essential determinants from irrelevant parameters, we can achieve inference with a fraction of the samples required by standard \textit{black-box} methods.

We implement this conceptual framework by leveraging the specific geometric structure of the residuals. We adopt a \textit{stochastic hierarchical} optimization algorithm informed by Sloppy Model theory \cite{Transtrum2010} that utilizes a reduced Hessian strategy. This operator functions as a linear autoencoder for the local landscape, efficiently identifying the \textit{stiff} (physically meaningful) parameter directions using minimal simulator queries. 
Furthermore, to ensure rigorous parameter inference in the presence of experimental and model errors, we construct the loss landscape using a principled probabilistic objective function. This formulation, derived from maximum likelihood estimation, explicitly accounts for uncertainties in the macroscopic observables, providing a robust metric for model optimization.

We demonstrate the efficacy of this framework in the field of plasma-surface interactions, a domain where these optimization challenges are particularly acute. The accurate description of plasma kinetics is essential to applications ranging from material processing \cite{Lieberman2005} and sustainable plasma catalysis \cite{Singh2025} to fusion energy \cite{Gianfranco2002}. 
However, the models describing these phenomena require surface reactivity parameters (e.g., sticking coefficients, energy barriers) that are often poorly defined or unknown \cite{Guerra2007, Afonso2024, Viegas2024}.
Calibrating these parameters is critical, yet the simulations are computationally expensive, rendering naive global search methods infeasible.

The specific case study addressed in this work is a mesoscopic surface kinetics model describing the recombination of atomic oxygen on a Pyrex wall in O$_2$/CO$_2$ glow discharge reactors, based on the framework established in \cite{Guerra2007, Afonso2024, Silveira2023, Blandine2024, Viegas2024}. To ensure a robust refinement, we utilize a comprehensive dataset comprising 225 distinct steady-state conditions compiled from the literature \cite{Viegas2024, Blandine2024, Silveira2023}. These experiments cover a wide operational space, including pressures from $0.2$ to $10$ Torr, discharge currents between $10$ and $40$ mA, and wall temperatures from $-20^{\circ}$C to $50^{\circ}$C.

To benchmark the performance of the proposed hierarchical approach, we compare it against five baseline algorithms selected to isolate different optimization challenges: Differential Evolution (DE) \cite{Storn1997} for global exploration; Covariance Matrix Adaptation Evolution Strategy (CMA-ES) \cite{hansen2001} for adaptation to ill-conditioned landscapes; Trust Region Reflective (TRF) \cite{Branch1999} and Powell's method \cite{powell1964} for local refinement; and Gaussian Processes (GP) \cite{Rasmussen2006} to evaluate the efficacy of statistical surrogate modeling.

The remainder of the paper is organized as follows: Section \ref{sec:Methods} presents the probabilistic problem formulation and the hierarchical algorithms explored. Section \ref{sec:Model_Simulator} details the experimental dataset and the physical description of the simulator model. Section \ref{sec:Results} presents the results of applying the proposed methods to the O$_2$ and CO$_2$ plasma-surface kinetics schemes. Finally, Section \ref{sec:Conclusion} discusses our findings and presents the conclusions.

\section{Methods}
\label{sec:Methods}

\subsection{Problem Formulation}

The foundation of the model is a surface kinetics scheme that describes the elementary processes occurring at the interface between a low-temperature plasma and a Pyrex (borosilicate glass) surface \cite{Guerra2007, Viegas2024, Afonso2024}. This scheme is translated into a system of coupled Ordinary Differential Equations (ODEs) that governs the temporal evolution of the chemical species' densities. The solution of this system depends on a vector of $n$ kinetic parameters, $\theta \in \mathbb{R}^n$, which includes terms such as energy barriers and steric factors of the reaction rate coefficients  \cite{Guerra2007}.

For a given set of parameters $\theta$, the system of ODEs is numerically integrated until an arbitrary time $\tau$ is reached. In this work, $\tau$ is assumed to be the duration required for the system to achieve stationarity.
From this $\tau$ state, a vector of $m$ macroscopic observables, $M(\theta) \in \mathbb{R}^m$, is computed. These observables may include quantities such as surface recombination probabilities \cite{Guerra2007, Viegas2024}, surface coverages, or gas species densities, which are directly compared to a set of experimental measurements, denoted by the vector $E \in \mathbb{R}^m$.
Hence, the problem is framed as an optimization exercise where the goal is to minimize the objective function:
\begin{equation}
    \theta^{*} = \arg \min_{\theta \in \Theta} \Phi(\theta),
    \label{eq:objective_problem_algos}
\end{equation}
where $\Theta \subseteq \mathbb{R}^{n}$ is the domain of physically admissible parameter vectors
(i.e., the set of all parameter vectors that satisfy the imposed physical constraints and limits), and $\Phi : \Theta \to \mathbb{R}_{\geq 0}$ is the objective loss given by:
\begin{equation}
    \Phi(\theta) =\frac{1}{2} \sum_{i \in\mathcal{D}} \sum^m_{j=1}\left(r^{(i)}_{j}(\theta) \right)^2,  ~~~~~~r^{(i)}_j(\theta) = \frac{E^{(i)}_{j} - M^{(i)}_{j}(\theta)}{E^{(i)}_{j}},
    \label{eq:objectve_function_algos}
\end{equation}
where we assume that the experimental measurements are independent and normally distributed. Moreover, $\mathcal{D}$ corresponds to the experimental dataset, and $r$ represents the residuals.
The full derivation of the objective function from the probabilistic framework is detailed in Appendix \ref{app:likelihood_derivation}.

\subsection{Baseline Optimization Algorithms}

We select five baseline algorithms, each representing a different strategy in non-linear least-squares optimization. These methods allow us to evaluate separately the challenges of global exploration, landscape adaptation, gradient estimation, local refinement, and surrogate modeling.

As a baseline for global exploration, we employ Differential Evolution (DE), a population-based heuristic method \cite{Storn1997}.
While effective at surveying the entire parameter space to avoid local minima, its performance is significantly hindered by the objective function's ill-conditioning, a well-known difficulty for the general class of sloppy models \cite{Transtrum2010} that we also analyze for our model in Section \ref{sec:Results}.
DE's use of isotropic search strategies struggles to make progress in landscapes with vastly different parameter sensitivities.

To specifically address the ill-conditioning difficulty, we use the Covariance Matrix Adaptation Evolution Strategy (CMA-ES) as an adaptive baseline \cite{hansen2001}. This method is more suited for the narrow, elongated valleys of our problem, as it learns the landscape's geometry by adapting its internal covariance matrix.

To benchmark standard local refinement strategies, we employ two distinct approaches. First, we include the Trust Region Reflective (TRF) algorithm \cite{Branch1999}, a robust bound-constrained extension of the classic Levenberg-Marquardt method \cite{More1978}, in which the Jacobian matrix is estimated via finite differences.
Second, we include Powell's method, a classic non-population-based algorithm \cite{powell1964}.
Unlike TRF, it operates by performing iterative line searches along a set of optimized directions without requiring any gradient estimation.

While our primary focus remains on methods that query the simulator directly, we also explore a widely used surrogate-based method, Gaussian Process (GP), for completeness \cite{Rasmussen2006}.
The implementation details of the baseline algorithms are presented in the Appendix \ref{app:Parameter_Details}.

\subsection{Hierachical Optimization}

\begin{figure}[h]
    \centering
    \includegraphics[width=0.6\linewidth]{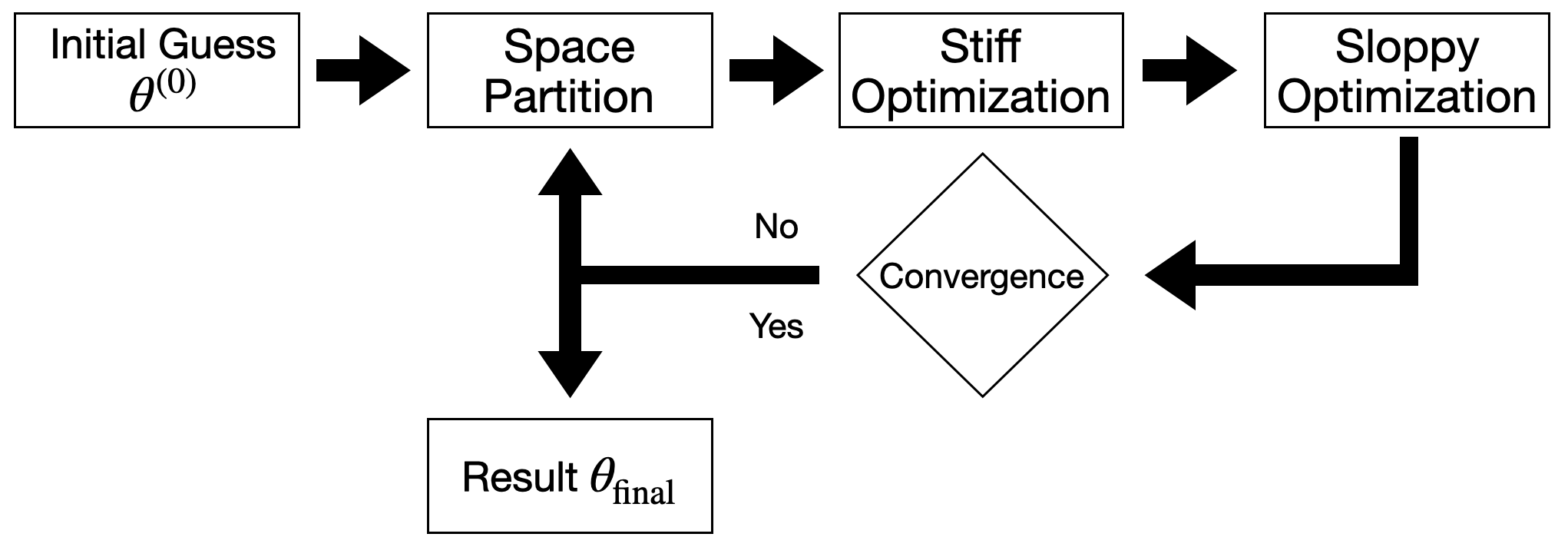}
    \caption{Hierarchical algorithm overview diagram}
    \label{fig:scheme_algo}
\end{figure}

Building on the insights that our loss $\Phi(\theta)$ is ill-conditioned and the difficulties found when exploring the baseline methods, we adopt an adaptive strategy that leverages the problem's geometric structure.
Informed by Sloppy Model theory  \cite{Transtrum2010, Transtrum_2011, Transtrum2012, Transtrum2015}, we reframe the search for a minimum as an iterative process guided by a geometric heuristic space partition \cite{transtrum2011efficient} that starts from a given initial parameter guess $\theta^{(0)}$.
The scheme \ref{fig:scheme_algo} presents a schematic overview of the algorithm considered. 

\subsubsection{Geometric Model Injection via the Gauss-Newton Hessian}

The heart of the heuristic space partition is a local model of the objective loss function $\Phi$ that explicitly takes into account the fact that the objective loss is dominated by a set of \textit{stiff} directions, while the remaining \textit{sloppy} modes contribute negligibly \cite{Transtrum2010, Transtrum_2011, Transtrum2012}.
This decomposition is based on the information provided by the Hessian matrix of the objective loss. Under the assumption that the residuals, $r$ (Eq.\eqref{eq:objectve_function_algos}), are small, the objective loss' Hessian of the estimate at iteration $i$, $\theta^{(i)}$, is well approximated by
\begin{equation}
[ H(\theta^{(i)})]_{mn} = [\nabla^2 \Phi(\theta^{(i)})]_{mn}\approx [H_{\rm GN} (\theta^{(i)})]_{mn} = \sum_l \frac{\partial r_l}{\partial \theta_m} \frac{\partial r_l}{\partial \theta_n},
\label{eq:Hessian_GN}
\end{equation}
where $H_{\rm GN}$ corresponds to the \textit{Gauss-Newton} approximation of the Hessian matrix \cite{nocedal2006}.
The eigendecomposition of this matrix, 
\begin{equation}
H_{\rm GN}(\theta^{(i)}) = V \Lambda V^T, ~~~\Lambda = \mathrm{diag} (\lambda_1, \dots, \lambda_n),
\label{eq:diag_H_GN}
\end{equation}
reveals a near-universal property of these models \cite{Transtrum2010, Transtrum2015}. As we will see in Fig. \ref{fig:panel1}(d), the eigenvalue spectrum $\{ \lambda_i \}$ decays exponentially, exposing a small set of \textit{stiff} parameter combinations that dominate the model's behavior and a large set of \textit{sloppy} combinations that have a negligible effect.
This justifies the hierarchical optimization approach. 
We partition the eigenpairs $\{ (\lambda_i, v_i) \}$ of $H_{\rm GN}$, where $v_i$ are the eigenvectors, into two different subspaces. 
The stiff subspace $V_s = \mathrm{span} \{ v_1, v_2, \dots, v_k \}$ is spanned by the eigenvectors corresponding to the large eigenvalues $(\lambda_i)$, while the \textit{sloppy} subspace, $V_l$, is its orthogonal complement. The splitting criteria is based on the cumulative variance conveyed by the eigenspectrum, analogous to Principal Component Analysis (PCA) \cite{Twarwat2016} (see Appendix \ref{app:hierarchical_algo} for details on the $\lambda_i$ and the subspace definitions). 

Any parameter vector, $\theta \in \mathbb{R}^n$, can be expressed as a deviation from $\theta^{(i)}$ along the natural coordinates defined by the eigenspace of $H_{\rm GN} (\theta^{(i)})$:
\begin{equation}
\theta = \theta^{(i)} + V^{(i)}_s \phi + V^{(i)}_l \psi.
\end{equation}
where $\phi \in \mathbb{R}^k$ and $\psi \in \mathbb{R}^{n-k}$ are the coordinate vectors representing the deviation within the stiff and sloppy subspaces, respectively.

Moreover, we further improve the sample efficiency in our optimization problem by reducing the dimensionality of the input space. 
By analyzing the eigenspectrum of $H_{\rm GN}$, we remove less significant dimensions, creating a \textit{reduced} sloppy subspace that focuses on its most important directions. Although this sacrifices a complete basis for the new coordinate system, our algorithm's iterative nature helps to mitigate this issue (see Appendix \ref{app:hierarchical_algo} for more details).

\subsubsection{Implicit Curvature via the Reduced Hessian Proxy}

While the full eigendecomposition (Eq. \ref{eq:diag_H_GN}) provides a complete map of the landscape, constructing the full Gauss-Newton Hessian requires computing the full  matrix $J_{mn} = \partial r_m / \partial \theta_n$. When the underlying simulator is computationally expensive, performing $n+1$ simulator evaluations to build $J$ becomes intractable for high-dimensional parameter spaces.

To circumvent this limitation, we employ a {reduced Hessian} strategy that acts as an efficient proxy to identify the stiff subspace. This approach can be viewed as a \textit{linear autoencoder} for the local curvature \cite{Transtrum2010}. Just as an autoencoder compresses data into a latent space to preserve the most significant features, we restrict our curvature search to a lower-dimensional random subspace $\Omega \in \mathbb{R}^{n \times k}$  (with subspace dimension $k\ll n$) to capture the maximum \textit{stiffness} energy \cite{halko2010}.

Mathematically, maximizing the curvature within this subspace (defined by the Rayleigh Quotient \cite{HornJohnson2012}) is equivalent to analyzing the eigenvalues of the reduced matrix $H_{\rm small}$:
\begin{equation}
H_{\rm small} = \Omega^{T} H_{\rm GN} \Omega = (J\Omega)^{T}(J\Omega).
\label{eq:reduced_hessian}
\end{equation}
This formulation allows us to compute the dominant curvature implicitly. By defining $Y = J\Omega$, we can approximate the columns of $Y$ directly via finite differences along the basis directions $\Omega$ \cite{Pearlmutter1994} (see Appendix \ref{app:hierarchical_deriv} for the derivation).
This reduces the computational cost from $n+1$ simulator calls to just $k+1$ calls, making it feasible to extract the stiff geometry even for expensive simulators.

The success of this approximation relies on the specific geometry of sloppy models. Provided the stiffness ratio (the gap between $\lambda_{\rm stiff}$ and $\lambda_{\rm sloppy}$) is sufficiently large, the energy landscape inside the random subspace is dominated by the projection of the true stiff modes \cite{Transtrum_2011}. Solving the eigenvalue problem for $H_{\rm small}$ yields the optimal rotation of our random basis, providing the approximate stiff basis $V_s$ required for the update rule in Eq. (3) (see Appendix \ref{app:random_projection_proof} for the derivation).

\subsubsection{Heuristic Meta-Simulator ($\mathcal{H}$)}

The space decomposition allows us to replace the original, intractable joint optimization with a more manageable, sequential one. We formalize this step as the application of the \textit{heuristic meta-simulator} operator, $\mathcal{H}_i$, where index $i$ signals that the operator is applied to the given estimate at iteration $i$, $\theta^{(i)}$. This operator maps the parameter vector $\theta^{(i)}$ to the next, $\theta^{(i+1)}$, by executing a search constrained by the geometric model derived at $\theta^{(i)}$. The operation $\theta^{(i+1)} = \mathcal{H}_i (\theta^{(i)})$ is a composition of three sequential subproblems:

\begin{itemize}
    \item {Stiff optimization:} We minimize the stiff space defined by $V_s$:
    \begin{equation}
        \theta^{(i')} = \theta^{(i)} + V^{(i)}_{s} \phi^*, \quad \text{where} \quad \phi^* = \arg \min_{\phi} \Phi(\theta^{(i)} + V^{(i)}_{s} \phi)
    \end{equation}
    
    \item {Sloppy re-alignment:} We re-orient the coordinate system to align with the local curvature of the sloppy subspace. We compute a reduced Hessian strictly within the sloppy subspace $V_l$ and solve its eigenproblem to obtain a rotation matrix $U$:
    \begin{equation}
        V'_{l} = V_{l} U
    \end{equation}
    This rotation decorrelates the parameters, ensuring that the subsequent optimizer moves along the valley's principal axes rather than traversing diagonally across ridges \cite{Transtrum_2011}.

    \item {Sloppy optimization:} We perform search along $V'_l$:
    \begin{equation}
        \theta^{(i+1)} = \theta^{(i')} + V'^{(i)}_{l} \psi^*, \quad \text{where} \quad \psi^* = \arg \min_{\psi} \Phi(\theta^{(i')} + V'^{(i)}_{l} \psi)
    \end{equation}
\end{itemize}

This sequential procedure constitutes a single step guided by the local geometry of the manifold.

\subsubsection{Iterative Refinement via Space Update}

A static model based on the geometry at $\theta^{(i)}$ is insufficient to navigate the objective surface $\Phi$. The key to the method's power is the \textit{iterative refinement} of the heuristic operator $\mathcal{H}_i$ in the style of \textit{Manifold Boundary Approximation Method} \cite{transtrum2011efficient}. After each step, the old operator $\mathcal{H}_i$ and its underlying basis $V_i$ are discarded. A new heuristic operator $\mathcal{H}_{i+1}$ is then constructed by re-evaluating the geometry on the new point $\theta^{(i+1)}$. Hence, the overall optimization at the iteration $i$, $\theta^{(i)}$, is a sequence of applications of these adaptive operators that starts from the initial guess $\theta^{(0)}$:
\begin{equation}
    \theta^{(i)} = ( \mathcal{H}_{i-1} \circ \dots \circ \mathcal{H}_{0}) (\theta^{(0)}),
\end{equation}
where $\circ$ denotes the composition operator.

This iterative process allows the algorithm to \textit{see} the changing landscape and adjust the coordinate system, enabling it to follow the curved, \textit{narrow valleys} characteristics of the \textit{sloppy} model. The formal description of the algorithm is presented in Appendices \ref{app:Parameter_Details} and \ref{app:hierarchical_algo}.

\subsubsection{Convergence and region of near-optimal parameters}
\label{sec:Analysis}

The algorithm guarantees reaching a stationary point. Indeed, since each subproblem is solved by a descent method, the objective function is guaranteed to be non-increasing: $\Phi(\theta^{(i+1)}) \leq \Phi(\theta^{(i)})$. Moreover, as $\Phi > 0$, the sequence must converge.

As termination criteria,  we consider the stability of the \textit{stiff} subspace \cite{transtrum2011efficient}. Convergence of the algorithm is assumed when the stiff subspace stops rotating, which is measured by analyzing the singular values $\{ \sigma_j \}$ of the singular value decomposition of the overlap matrix $V^{(i+1), T}_s V^{(i)}_s$. The misalignment, $\delta_s = \max_j (1-\sigma_j)$, indicates convergence when it falls below a tolerance, $\delta_s < \varepsilon_{\rm stop}$.

The criterion ensures the converged point, $\theta_{\rm final}$, has \textit{almost} the properties of a true minimum, as it is ensured  that the final gradient lies in the \textit{sloppy} subspace, which implies that the $l_2$-norm of the gradient is bounded:
\begin{equation}
	\| \nabla \Phi(\theta_{\rm final}) \|_2 \leq  |\lambda_{\rm max ~sloppy}| \cdot \| \Delta \theta \|_2.
    \label{eq:Upper_bound_gradient}
\end{equation}
where $\Delta \theta$ is the difference between $\theta_{\rm final}$ (the algorithm's estimation at the last iteration) and the true local minimum $\theta_{\rm min}$. Appendix \ref{subsec:Upper_bound_Gradient} presents the full derivation of Eq.\eqref{eq:Upper_bound_gradient}.
Since $\lambda_{\rm max ~ sloppy}$, the largest eigenvalue (in magnitude) of the \textit{sloppy} space is small ($ \lambda_{\rm max ~ sloppy} / \lambda_{\rm max ~ stiff} \ll 1$), the norm of the gradient at the converged point is guaranteed to be small.

The algorithm is formally a local method and does not guarantee finding the global minimum, $\theta_{\rm global}$. However, this procedure is often sufficient for sloppy models, where the set of near-optimal parameters, $\mathcal{S}_{\epsilon} = \{ \theta \in \mathbb{R}^n \mid \Phi(\theta) \leq \Phi(\theta_{\rm global}) + \epsilon \} $ with $\epsilon > 0$, is typically a large, high-dimensional, and extended region in the parameter space \cite{Transtrum2010, Transtrum_2011, Transtrum2012}.

The reason for the sloppy structure can be understood by analyzing the geometry of the loss function $\Phi$ around a found minimum, $\theta_{\rm min}$. The Hessian evaluated at this point, $H \equiv H(\theta_{\rm min})$, which has eigenvectors $\{ v_i\}$ and eigenvalues $\{\lambda_i \}$. The change in the loss function, $\Delta \Phi$, for small deviation $\Delta \theta$ is given by the quadratic approximation:
\begin{equation} 
\Delta \Phi \approx \frac{1}{2} \Delta \theta^T H \Delta \theta .
\end{equation}
Assuming that deviations $\Delta \theta$ lie solely within the sloppy subspace, which is spanned by the eigenvectors $\{ v_j \}_{j \in \rm sloppy}$, the deviations can be written as $\Delta \theta = \sum_{j \in \rm slopppy} \alpha_j v_j$. 
Replacing this in the quadratic approximation gives:
\begin{equation} 
\Delta \Phi \approx \frac{1}{2} \left( \sum_{i \in \mathrm{sloppy}} \alpha_i v_i^T \right) H \left( \sum_{k \in \mathrm{sloppy}} \alpha_k v_k \right) 
\end{equation}
By expressing the Hessian in its eigenbasis, $H = \sum_j \lambda_j v_j v_j^T$, and using the orthonormality of the eigenvectors $(v^T_i v_j = \delta_{ij})$, it simplifies to:
\begin{equation} 
    \Delta \Phi \approx \frac{1}{2} \sum_{j \in \mathrm{sloppy}} \lambda_j \alpha_j^2 
\end{equation}
Since all sloppy eigenvalues $\lambda_j$ are small (many orders of magnitude smaller than the stiff eigenvalues), this equation shows that large displacements in parameter space along the sloppy space (large coefficients $\alpha_j$) produce negligible changes in $\Phi$. This 
explains the extended region $\mathcal{S}_{\epsilon}$.
Conversely, any deviation along a stiff direction would be heavily penalized by its large eigenvalue, confining the set of good parameters \cite{Transtrum2015}.

\section{Physical System and Experimental Data}
\label{sec:Model_Simulator}

\subsection{Simulator's Physical Model: Surface Kinetics on Pyrex}

The case study analyzed in this work concerns a mesoscopic surface kinetics model that describes
the interaction between an O$_2$/CO$_2$ glow discharge and a Pyrex reactor wall. 
To the best of our knowledge,
this formulation represents the current state-of-the-art for describing plasma-surface recombination.
The model is developed within the theoretical framework established in earlier studies \cite{Guerra2007, Silveira2023, Viegas2024, Afonso2024, Blandine2024}, and both encompasses and extends the earlier results from \cite{Guerra2007, Viegas2024, Afonso2024}, which investigated O-atom recombination in O$_2$ glow discharges, as well as the preliminary findings for CO$_2$ plasmas reported in \cite{Silveira2023,Blandine2024}.
The model accounts for adsorption, surface diffusion, and recombination of key reactive species, including atomic oxygen (O), molecular oxygen (O$_2$), ozone (O$_3$),  carbon monoxide (CO), and carbon dioxide (CO$_2$).

The model considers three distinct adsorption sites on the Pyrex surface. Physisorption sites (\ce{F_V}) cover the entire surface, binding particles through weak, reversible van der Waals forces. Chemisorption sites (\ce{S_V}), located at surface defects, form strong, irreversible chemical bonds. Additionally, at low pressures ($< 1$  Torr), ion bombardment can create highly reactive, reversible metastable sites (\ce{S^*_V}). Recombination of atomic species occurs via two main mechanisms: the Eley-Rideal (E-R) mechanism, where the gas-phase particle, \ce{A(g)},  reacts directly with an adsorbed particle \ce{A_{(F/S)}} (F in the case of a physisorption site and S for the chemisorption one) \hspace{2.5 cm} (\ce{A(g) + A_{(F/S)} -> A_2(g) + (F/S)_V}) and the Langmuir-Hinshelwood (L-H) mechanism, involving two adsorbed particles 
(\ce{A_F + A_{(F/S)} -> A_2(g) + (F/S)_V + F_V}).
The full reaction scheme comprises 62 elementary reactions and a total of 125 parameters \cite{Viegas2024, Blandine2024}.
The oxygen atom kinetics is temperature-dependent: at low pressures ($T_w < 25^{\circ}$C), recombination is dominated by reactions like (\ce{O_F + O_F -> O_2(g) + 2 F_V}) (L-H), while at higher temperatures ($T_w > 25^{\circ}$C), reactions involving chemisorbed oxygen, such as (\ce{O(g) + O_S -> O_2(g) + S_V}) (E-R), become dominant. The inclusion of carbon species is critical for CO$_2$ plasmas, where the primary effect is the passivation of reactive chemisorption sites by CO, which competes with oxygen and significantly reduces the overall O-atom recombination rate \cite{Blandine2024, Silveira2023, morillocandas}.  
The net effect of this micro-kinetic scheme is consolidated into a single macroscopic observable $M$: the effective recombination probability of atomic oxygen, denoted as $\gamma_{\rm O}$.  Crucially, while limited to a single scalar, this observable encodes complex dependencies on pressure, discharge current, and gas mixture, measured across 225 different experimental conditions.

\subsection{Parameters for Optimization}

The predictive accuracy of the kinetic model depends on its rate constants, many of which are not well understood from first principles or experimental data. 
The goal of this work is to refine a subset of the model parameters, namely the Arrhenius parameters ($k_{0,i}$ and $E_{a,i}$), by fitting the model macroscopic observable $\gamma_{\rm O}$ to experimental data. These parameters define the rate coefficients, $k_i = k_{0,i} \cdot \exp{(-E_{a\,i} / k_B T)}$, that are used within the model \cite{Guerra2007}.
The parameters to be optimized are presented in a vector $\theta$, which includes:
\begin{itemize}
    \item Energy barriers ($E_{a \, i}$): the activation energies for the creation and destruction of metastable surface sites (\ce{S^*_V}) 
    \item Steric Factors ($k_{0,i}$) : The pre-exponential factors for the chemical equations involving CO and the metastable sites.
    \item Desorption Frequency Parameters: For the desorption of physisorbed species (\ce{O_F, O2_F, CO_F}), the desorption frequency is described by the parameterization $\nu_d(T) = 10^{15} \times \left( A + B \cdot \exp{(E / k_B T)} \right)$ s$^{-1}$ \cite{Viegas2024}, where the coefficients $A, B$, and $E$ are to be optimized.
\end{itemize}

In total, the optimization problem addresses 29 distinct parameters, which correspond to the ones with the highest uncertainty estimation \cite{Viegas2024, Blandine2024} 

\subsection{Experimental Validation Dataset}

The dataset, compiled from the literature \cite{Viegas2024} and from unpublished measurements used in \cite{Blandine2024, Silveira2023}, provides the $\gamma_{\rm O}$ observable across 225 distinct steady-state conditions. These experiments cover a wide operational space, including \ce{O_2}/\ce{CO_2} gas mixtures with a total flow rate of $7.4$ sccm, pressures from $0.2$ to $10$ Torr, discharge currents between $10$ and $40$ mA, and wall temperatures from $-20^{\circ}$C to $50^{\circ}$C.

\section{Results}
\label{sec:Results}

In this section, we evaluate the performance of the proposed hierarchical algorithm against the baseline methods. To address validation concerns and demonstrate predictive capability beyond simple curve fitting, we partition the experimental dataset into a {training set (80\%, $N=180$)} and a held-out {test set (20\%, $N=45$)}. 
All optimizations were restricted to the training set, while the test set was used post-optimization to verify the generalization of the found solutions. As required for a fair comparison, all algorithms were initialized with the same parameter guess, $\theta^{(0)}$, which significantly differs from the literature default ($\Phi(\theta^{(0)}) \approx 700$ vs. $\Phi(\theta_{\rm default}) \approx 0.1$).

\begin{figure}[h]
    \centering
    \includegraphics[width=1.0\linewidth]{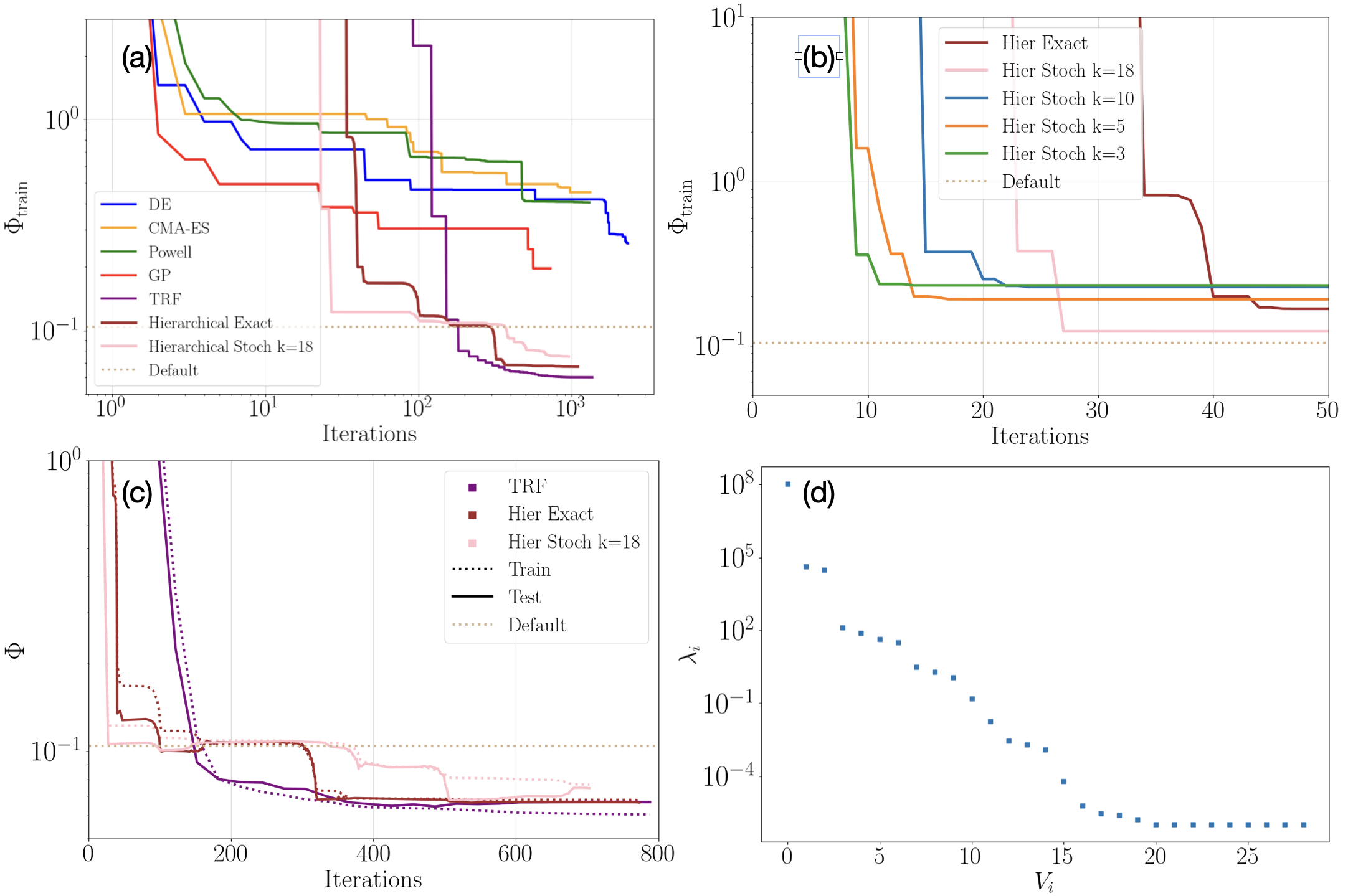}
\caption{
    Optimization performance and geometric analysis.
    Figure (a) compares the convergence of the training loss ($\Phi_{\rm train}$) versus simulator iterations for the hierarchical methods (brown and pink) against the baseline algorithms, where the dotted line marks the default parameter loss.
    Figure (b) presents the sample efficiency in the early optimization phase (first 50 iterations) for the Hierarchical Stochastic strategy using varying subspace sizes $k$.
    Figure (c) displays the simultaneous evolution of the training loss (dotted lines) and test loss (solid lines) for the best-performing algorithms (Hierarchical and TRF).
    Figure (d) shows the eigenvalue spectrum ($ \{ \lambda_i \} $) of the Gauss-Newton Hessian, $H_{\rm GN}$, as a function of the eigendirections $V_i$ at the minimum $\theta_{\rm final}$ found. The $H_{\rm GN}$ is represented as $H_{\rm GN} = V \Lambda V^T$, with $\Lambda= \mathrm{diag}(\lambda_1, \dots,\lambda_n)$. In this case, $V_0$ corresponds to the stiff subspace, while the remaining directions $V_i$ are the sloppy subspace.
    }
    \label{fig:panel1}
\end{figure}

Figure \ref{fig:panel1}(a) presents the global convergence of the training loss ($\Phi_{\rm train}$) for all algorithms. 
Critically, the x-axis represents cumulative simulator calls. Since the kinetic simulation constitutes the dominant computational bottleneck, this axis serves as the direct metric for sample efficiency.
The Hierarchical methods, comprising the exact full-Jacobian formulation (red) and the stochastic proxy with k=18 (pink), demonstrate superior convergence speed compared to the global (DE, CMA-ES) and gradient-free local (Powell) baselines. While the Trust Region Reflective algorithm (TRF, purple line) proves to be a strong competitor, eventually reaching a similar loss floor, the Hierarchical approach achieves high-quality solutions with significantly fewer iterations in the intermediate regime. This acceleration is driven by the algorithm's ability to exploit the highly anisotropic structure of the loss landscape, rapidly converging onto the low-energy manifold (the valley floor) before refining the sloppy parameters.

Crucially, for computationally expensive simulators, the primary metric of interest is sample efficiency: the ability to find good parameters with minimal model evaluations. Figure \ref{fig:panel1}(b) analyzes this by comparing the \textit{Hierarchical Exact} strategy against the \textit{Hierarchical Stochastic} (reduced Hessian) strategy with varying subspace sizes $k$. The results reveal a striking advantage for the stochastic approach: even with subspace dimensions as low as $k=3$ or $k=5$, a massive compression relative to the full parameter space of $n=29$, the algorithm achieves a rapid loss reduction in the first 20 iterations, significantly outperforming the Exact method. This confirms that the stiffest features of the landscape can be identified and exploited using only a coarse, low-rank approximation of the curvature, validating the framework's scalability to more complex models where $n$ is large.

To ensure that these optimized parameters correspond to a physically meaningful model rather than an artifact of overfitting, we examine the generalization error in Figure \ref{fig:panel1}(c). Here, we track both the training loss (dotted lines) and test loss (solid lines) throughout the optimization for the top-performing algorithms (TRF and Hierarchical). The close alignment between the training and test trajectories indicates that the algorithms are not overfitting. The Hierarchical method achieves a final test loss comparable to the TRF baseline, confirming that the \textit{sloppy} geometric separation is a robust feature of the physical model that allows for reliable extrapolation to unseen experimental conditions.

Finally, Figure \ref{fig:panel1}(d) displays the eigenvalue spectrum of the Hessian at the found minimum. The rapid decay of eigenvalues ($\lambda_i$) spanning several orders of magnitude confirms the underlying sloppy nature of the model, justifying the hierarchical separation of parameter space that drives our algorithm's efficiency.

To assess the robustness of the identified parameters and ensure they represent physical reality rather than statistical artifacts, the optimization was executed using the {Hierarchical Exact} algorithm, starting from the previous initial guess ($\Phi(\theta^{(0)}) \approx 700$). This procedure was performed independently on five distinct training/test splits (80\%/20\%), where we employed stratified random sampling based on the joint Pressure-Temperature distribution. This ensures that both the training ($80\%$) and test ($20\%$) sets uniformly cover the experimental operating window, preventing the clustering of specific thermodynamic conditions in a single subset.

\begin{figure}[H]
    \centering
    \includegraphics[width=1.0\linewidth]{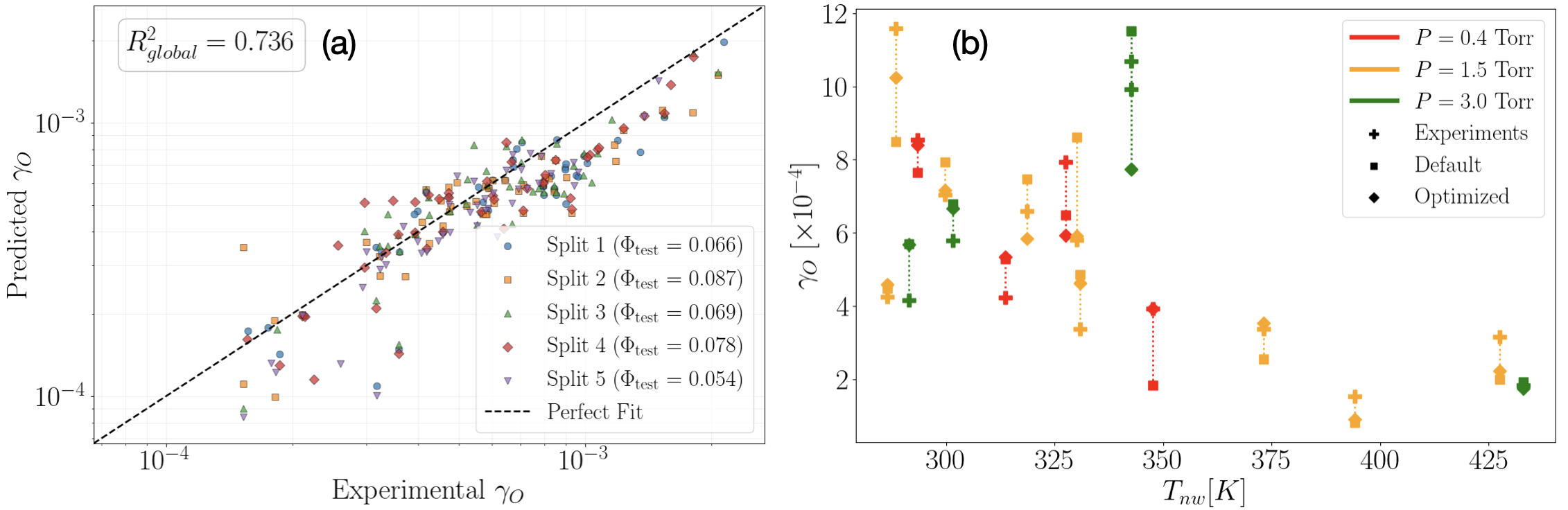}
    \caption{Validation of the optimized kinetics model. (a) Parity plot of predicted vs. experimental $\gamma_O$ for the test sets of all 5 independent splits, showing a strong correlation ($R^2=0.736$). (b) Detailed comparison of $\gamma_O$ as a function of wall temperature for selected test conditions at three different pressures, highlighting the accuracy gain of the optimized parameters (diamonds) over the default values (squares).}
    \label{fig:validation_panel}
\end{figure}

The global predictive capability of the optimized model is validated first in Figure \ref{fig:validation_panel}(a), which presents a parity plot of the predicted versus experimental $\gamma_O$ probabilities for all test set conditions across the five splits. The model achieves a global $R^2$ of $0.736$, with consistent test loss values ($\Phi_{\rm test}$) across all splits (ranging from 0.054 to 0.087). This uniformity demonstrates that the algorithm has found a generalizable physical solution rather than overfitting to specific training subsets.

Figure \ref{fig:validation_panel}(b) further details the model's performance by comparing simulations against experimental measurements for specific test conditions at varying pressures ($p=0.4, 1.5, 3.0$ Torr). The optimized parameters (diamonds) yield a closer agreement with the experimental data (crosses) compared to the default parameters (squares), particularly in reproducing the temperature dependence of $\gamma_{\rm O}$ observed in the experiments. This improvement validates the hierarchical optimization approach as a robust tool for refining surface kinetics models in regimes where experimental data is sparse or noisy.

\begin{figure}[H]
    \centering
    \includegraphics[width=0.9\linewidth]{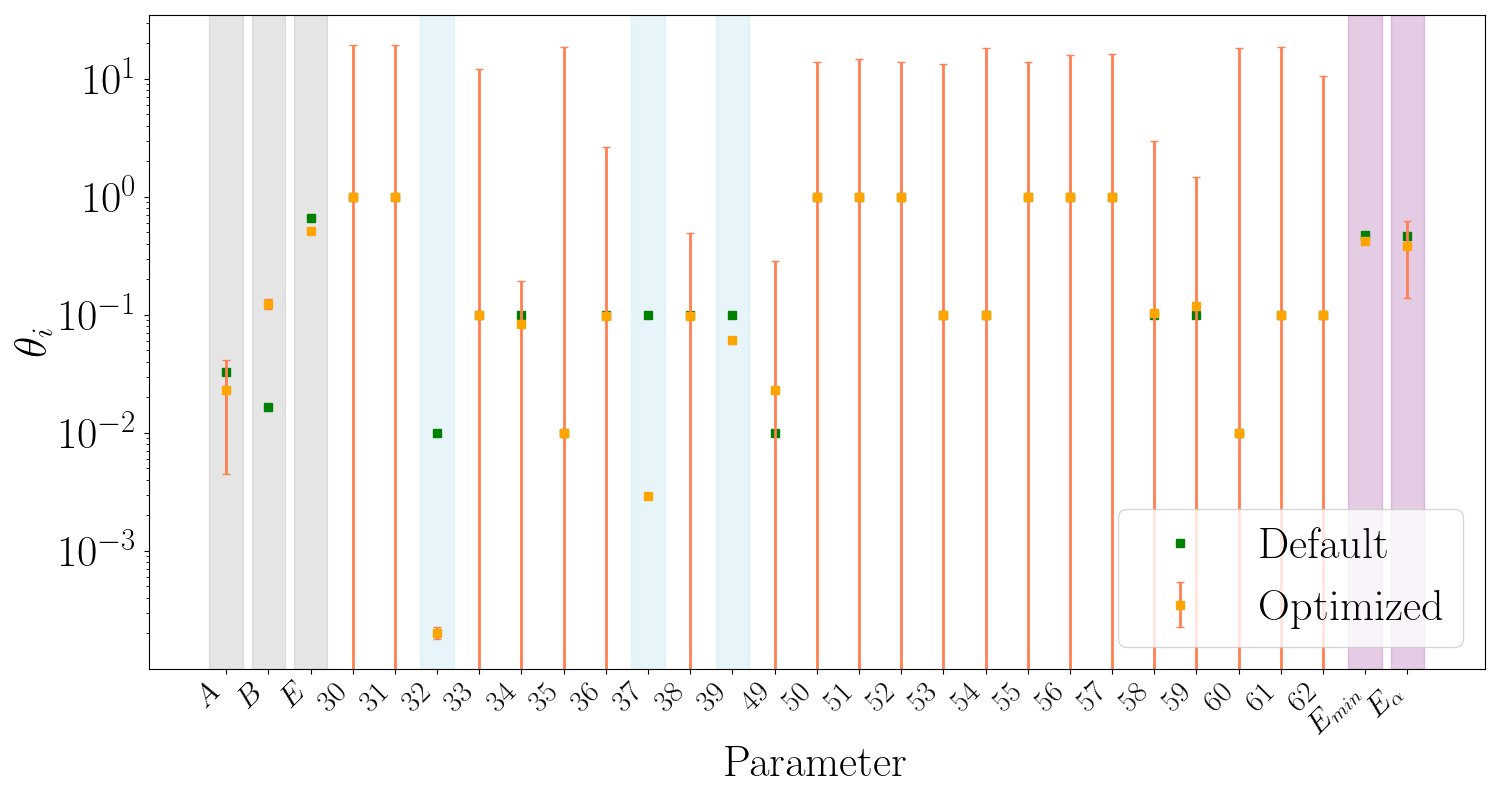}
    \caption{Comparison between the normalized default literature parameters (green) and the median optimized values (orange) obtained from 5-fold cross-validation. 
    The numeric labels correspond to the steric factor indices (e.g., the number 30 corresponds to $k_{0,30}$, following the sequential reaction order of the full kinetic scheme.
    The error bars correspond to the Hessian-derived uncertainty bounds. Colored bars highlight the stiff parameters (Gray: Desorption; Blue: CO chemisorption; Pink: Metastable energies), while uncolored bars indicate sloppy parameters.}
    \label{fig:params_uncertainty}
\end{figure}

With the model's predictive power established, we examine the stability of the physical parameters themselves. Figure \ref{fig:params_uncertainty} summarizes the results, comparing the literature default parameters (green) with the median optimized values (orange) obtained across the five independent runs. The error bars represent the uncertainty quantified via the Hessian of the {training loss} calculated on the corresponding split at the median solution. The intervals are given by $\Delta \theta_i = \sqrt{2 \, \Delta \Phi \cdot (H^{-1}_{\rm GN})_{ii}} \,$ \cite{Kalmikov2014}, with a confidence interval threshold of $\Delta \Phi = 0.01 \times \Phi(\theta_{\rm final})$.

The magnitude of these error bars effectively maps the \textit{stiffness} of the physical model: small bars indicate stiff parameters that are tightly constrained by the physics, while large bars reveal sloppy parameters where the model lacks sensitivity. As detailed in Table \ref{tab:stiff_params}, the critical (stiff) parameters governing the system are identified as:
\begin{itemize}
    \item The desorption frequencies of \ce{O_F}, \ce{O2_F}, and \ce{CO_F} (parameters $A$, $B$, and $E$, highlighted in gray);
    \item The steric factors for the key CO chemisorption lifecycle reactions: its formation ($k_{0, 32}: \, \ce{CO(g) + S_V -> CO_S}$) and its recombination with atomic oxygen ($k_{0, 37}: \,\ce{O(g) + CO_S -> CO_2 + S_V}$ and $k_{0, 39}: \,\ce{O_F + CO_S -> CO_2(g) + S_V}$) (highlighted in blue);
    \item The characteristic energies for the creation and destruction of metastable states: $E_{\alpha}$ and $E_{\min}$ (highlighted in pink).
\end{itemize}

Conversely, the steric factors for the remaining metastable rate coefficients and reactions involving physisorbed CO adsorption/desorption remain poorly constrained, as illustrated by the large uncolored error bars in Figure \ref{fig:params_uncertainty}. This confirms that their broad uncertainties are a structural consequence of the model's sloppiness rather than insufficient data.

\begin{table}[htbp]
    \centering
    \begin{tabular}{ccc} 
        \toprule
        \textbf{Parameter} & \textbf{Default} & \textbf{Optimized (Median)} \\
        \midrule
        A & $1.634\times 10^{-2}$ & $1.158\times 10^{-2}$ \\
        B   & $1.670\times 10^{-2}$ & $1.245\times 10^{-3}$ \\
        E [kJ/mol] & 19.75 & 15.27 \\
        $k_{0,32}$ & $10^{-2}$ & $2.9\times10^{-3}$ \\
        $k_{0,37}$ & $10^{-1}$ & $2.2\times10^{-3}$ \\
        $k_{0,39}$ & $10^{-1}$ & $6.1 \times 10^{-2}$ \\
        E$_{\rm min}$ [eV] & 2.90 & 2.69 \\
        E$_{\alpha}$ [eV] & 0.29 & 0.27 \\
        \bottomrule
    \end{tabular}
    \caption{Default and optimized values (median of 5 splits) for the stiff parameters.}
    \label{tab:stiff_params} 
\end{table}

\section{Conclusion and Outlooks}
\label{sec:Conclusion}

This work presents a novel data-driven framework for the optimization of  physical models, addressing the fundamental challenge of learning from sparse data in computationally expensive systems. By bridging the gap between physical modeling and Scientific Machine Learning, we depart from standard black-box optimization strategies to introduce a {hierarchical} procedure that actively exploits the intrinsic geometry of the parameter space.

Central to our approach is the recognition that many physical systems are {sloppy}: their behavior is dominated by a low-dimensional latent manifold rather than the full parameter set. We leverage this structure by decomposing the optimization into {stiff} and {sloppy} subspaces. This geometric strategy allows the algorithm to rapidly converge onto the low-energy manifold, the \textit{valley floor}, before refining the parameters along the softer directions, effectively mitigating the severe ill-conditioning that paralyzes traditional baseline optimizers.

A key methodological contribution is the implementation of this strategy via a {stochastic reduced Hessian}. We demonstrated that this low-rank approximation functions effectively as a linear autoencoder for the local curvature, extracting the latent structure of the landscape using a constant number of evaluations proportional to the intrinsic dimension ($k$) rather than the full parameter count ($n$). This property ensures the method scales favorably for high-dimensional problems and remains robust in low-data regimes.

We validated this framework by applying it to the refinement of mesoscopic surface kinetic schemes, which are essential for describing plasma-surface interactions. Specifically, we addressed the interaction of O$_2$ and CO$_2$ glow discharges with Pyrex reactor walls, utilizing a comprehensive dataset of 225 steady-state experimental conditions.  Beyond optimization, we utilized the Hessian to derive data-driven uncertainty estimates. This analysis quantitatively confirmed that parameters related to the desorption frequencies of oxygen and carbon monoxide, the steric factors controlling chemisorbed CO formation and oxidation, and the energies required to create and destroy metastable states are well-constrained by the experimental data. In contrast, the other parameters are less sensitive.

Future research will focus on two distinct directions. First, we aim to analyze the insights gained from this data-driven approach to explore the underlying physics of atomic oxygen recombination on Pyrex surfaces. 
Second, while the current work focused on robust interpolation within the observed operational space, investigating the model's limits in extrapolation remains a compelling challenge. 

The proposed framework offers a model-agnostic solution that maximizes information extraction by targeting the problem's geometric stiffness. It effectively bridges the gap between high-dimensional physical complexity and the constraints of data scarcity. Consequently, this approach establishes a transferable methodology for addressing inverse problems, extending well beyond the specific context of plasma kinetics.

\section*{Acknowledgments}
The authors gratefully acknowledge Tiago Dias, Ana Sofía Morillo Candás, and Olivier Guaitella for providing unpublished experimental measurements obtained at \textit{Laboratoire de Physique des Plasmas}, which were used in the experimental dataset for this work.

\section*{Funding}
This work was supported by the Portuguese FCT - Fundação para a Ciência e a Tecnologia, under project's references UID/50010/2023, UID/PRR/50010/2025 and LA/P/0061/2020. This work was further supported by the European Union under Horizon Europe project CANMILK (DOI:10.3030/101069491).
PV acknowledges support by project CEECIND/00025/2022 of FCT.

% This section is a list of funder names and grant numbers

% \roles{Sample text inserted for demonstration.}
% List author names and the contributions made to the article, using terms from the NISO Contributor Roles Taxonomy (CRediT) https://credit.niso.org

\section*{Data availability}
The code used to produce the results and figures in this paper is publicly available on GitHub at \hyperlink{https://github.com/joseAf28/PlasmaDM}{https://github.com/joseAf28/PlasmaDM}. The data supporting the findings of this study is available within the repository.

\nocite{*}

%Bibliography
\bibliographystyle{unsrt}  
\bibliography{references}

\section*{Appendix}
\appendix

\section{Observed-data likelihood derivation}
\label{app:likelihood_derivation}

Here, we present the detailed derivation of the objective loss function based on a latent variable model.
We assume the system is governed by an unobserved (latent) true state $x^*$,  which is deterministically predicted by the model parameters $\theta$. The observables, $M = \{ \tilde{M}, x^{\rm exp}\}$, are treated as noisy measurements conditioned on this latent state. The observables are divided into two groups based on their dependencies:
\begin{itemize}
    \item $\tilde{M}$ represents measurements that are function of both the latent state $x^*$ and the hyperparameters $\theta$ directly (such as the recombination probability)
    \item $x^{\rm exp}$ represents measurements that depend only on the latent space  (such as surface or gas phase densities).
\end{itemize}

In the specific case of the physical system studied, the macroscopic observable corresponds to the recombination probability, $\gamma_O$, which is of the type $\tilde{M}$. This indicates that for this physical application $M=\{ \gamma_O \}$.

\subsection{Likelihood Formulation}

The joint likelihood of the observed data for a single experiment is obtained by marginalizing over the latent state $x^*$:
\begin{equation}
p(\tilde{M},x^{\rm exp} | \theta) = \int dx^*  \ p(\tilde{M} | x^*,  \theta) \ p(x^{\rm exp} | x^*) \ p(x^* | \theta).
\label{eq:general_prob_model}
\end{equation}

The core assumption is that the latent state is a deterministic function of the parameters, which we denote as $x(\theta)$. This is expressed as a Dirac delta distribution: $p(x^*|\theta) = \delta(x^* - x(\theta))$. Substituting this into Eq.\eqref{eq:general_prob_model} collapses the integral to:
\begin{equation}
p(\tilde{M},x^{\rm exp} | \theta) =  p(\tilde{M} | x(\theta), \theta) \ p(x^{\rm exp} | x(\theta)).
\label{eq:deterministic_likelihood}
\end{equation}
We assume that all measurements are corrupted by additive Gaussian noise, such that:
\begin{align}
p(x^{\rm exp}|x(\theta)) &= \mathcal{N}(x^{\rm exp} | x(\theta), \Sigma_x), \\
p(\tilde{M} | x(\theta), \theta) &= \mathcal{N}(\tilde{M}|\tilde{M}(x(\theta), \theta), \Sigma_M),
\end{align}
where $\tilde{M}(x(\theta), \theta)$ is the model prediction for the macroscopic observables $\tilde{M}$, and $\Sigma_x$ and $\Sigma_{\tilde{M}}$ are the covariance matrices.

\subsection{Approximation for the Loss Function}

The likelihood in Eq.\eqref{eq:deterministic_likelihood} depends on the unobserved latent state $x(\theta)$. To create a loss function based only on the observed data $x^{\rm exp}$, we linearize the model prediction $\tilde{M}(x(\theta),\theta)$ around $x^{\rm exp}$:
\begin{equation}
\tilde{M}(x(\theta), \theta) \approx \tilde{M}(x^{\rm exp}, \theta) + J^{(\tilde{M})}_x(\theta) (x(\theta) - x^{\rm exp}),
\end{equation}
where $J_x^{(\tilde{M})}(\theta)$ is the Jacobian of $\tilde{M}$ with respect to $x$ and $\delta x = x(\theta)-x^{\rm exp}$. From our assumption for $p(x^{exp} | x(\theta))$, the random variable $\delta x$ follows a Gaussian distribution with zero mean, $\mathbb{E}[\delta x]=0$, and covariance $\Sigma_x$.

The distribution for $\tilde{M}$ conditioned on the observed $x^{\rm exp}$ can now be approximated as a new Gaussian, $\tilde{p}(\tilde{M}|x^{\rm exp}, \theta) = \mathcal{N}(\mu(\theta), \tilde{\Sigma}_{\tilde{M}}(\theta))$, whose mean and covariance parameters are given by:
\begin{align}
\mu(\theta) &= \tilde{M}(x^{\rm exp}, \theta) + J^{(\tilde{M})}_x(\theta) \,\mathbb{E}[\delta x] = \tilde{M}(x^{\rm exp}, \theta), \\
\tilde{\Sigma}_{\tilde{M}}(\theta) &= \Sigma_{\tilde{M}} + J^{(\tilde{M})}_x(\theta) \text{Cov}(\delta x) J_x^{(\tilde{M})\, T}(\theta) = \Sigma_{\tilde{M}} + J^{(\tilde{M})}_x(\theta) \Sigma_x J_x^{(\tilde{M})\, T}(\theta).
\end{align}

The total negative log-likelihood for a dataset
$\mathcal{D} = \{ (x^{\rm exp}_i, \tilde{M}_i)\}^D_{i=1}$ is the objective loss function $\mathcal{L}(\theta) = \sum_{i=1}^D \mathcal{L}_i(\theta)$. For a single data point, this is:
\begin{equation}
\begin{split}
\mathcal{L}_i(\theta) &= - \log \tilde{p}(\tilde{M}_i | x^{\rm exp}_i, \theta) - \log p(x_i^{\rm exp} | x_i(\theta)) = \\
&= \frac{1}{2} (\tilde{M}_i - \mu_i(\theta) )^T \tilde{\Sigma}^{-1}_{\tilde{M},i}(\theta) (\tilde{M}_i - \mu_i(\theta) ) + \frac{1}{2} \log | 2\pi\tilde{\Sigma}_{\tilde{M},i}(\theta) | \\
&\quad + \frac{1}{2} (x^{\rm exp}_i - x_i(\theta) )^T \Sigma^{-1}_{x,i} (x^{\rm exp}_i - x_i(\theta) ) + \frac{1}{2} \log | 2\pi\Sigma_{x,i} |.
\end{split}
\end{equation}

\subsection{Simplified Loss for Relative Errors}
We can obtain a simpler, unweighted loss function under several assumptions:

\begin{enumerate}
\item The model output $M$ is insensitive to $x$, so the Jacobian $J^{(\tilde{M})}_x(\theta) \approx 0$. This implies $\mu_i(\theta) \approx \tilde{M}(x^{\rm exp}_i, \theta)$ and $\tilde{\Sigma}_{\tilde{M},i} \approx \Sigma_{\tilde{M},i}$. 
\item The covariance matrices $\Sigma_{\tilde{M}}$ and $\Sigma_x$ are diagonal and constant with respect to $\theta$. This allows us to drop the $\log | \cdot |$ terms. 
\item The measurement error is proportional to the measurement value. For the diagonal covariances, this means $\Sigma_{\tilde{M},ii} \propto (\tilde{M}^{\rm exp}_i)^2$ and $\Sigma_{x,ii} \propto (x^{\rm exp}_i)^2$.
\end{enumerate}

Under these assumptions, and ignoring constant scaling factors, the loss simplifies to a sum of squared relative errors:
\begin{equation}
\Phi(\theta) = \frac{1}{2} \sum^D_{i=1} \left[ \left \lVert \frac{\tilde{M}^{\rm exp}_i - \tilde{M}(x^{\rm exp}_i, \theta)}{\tilde{M}^{\rm exp}_i} \right\rVert^2_2 + \left \lVert \frac{x^{\rm exp}_i - x_i(\theta)}{x^{\rm exp}_i} \right\rVert^2_2 \right], 
\end{equation}
where the division is element-wise and $\lVert \cdot \rVert_2$ is the $l_2$-norm. Noting that $M = \{ \tilde{M}, x^{\rm exp}\}$ and unrolling the $\|\cdot\|^2_2$ into its explicit sum, we obtain Eq.\eqref{eq:objectve_function_algos}.

\section{Optimizer Implementation Details}
\label{app:Parameter_Details}

All optimizations were performed on parameters normalized to the unit interval $[0,1]$.

\subsection{Baseline Algorithms}

\begin{itemize}
    \item Differential Evolution (DE): We used \textit{scipy.optimize.differential$\_$}evolution \cite{Virtanen2020} with \textit{best1bin} strategy, \textit{recombination}$=0.7$, \textit{tol}$=0.1$, \textit{polish}=True and \textit{seed}=42. The population size and maximum generations were set to $15$ and $90$, respectively;

    \item CMA-ES:  We used the \textit{cma} library \cite{hansen2016}. \textit{CMAEvolutionStrategy} was initialized with $\sigma_0=1.0$, \textit{maxiter}$=70$, \textit{tolx}$=10^{-6}$, and a population size of $4+4\cdot \log(n)$ (where $n$ is the parameter dimension);

    \item Powell (Baseline): We optimized the full parameter space using only Powell's method via \textit{scipy.optimize.minimize(method="Powell")} \cite{Virtanen2020}, with its default termination criteria

    \item GP Optimization: We also optimized the normalized parameter space using Gaussian Process Bayesian optimization via \textit{skopt.gp\_minimize} \cite{gp_optimize}. The search was initialized with 40 random points, seeded with the default parameters, and run for a total of 980 function evaluations.

    \item  Trust Region Reflective (TRF): We used \textit{scipy.optimize.least\_squares} with \textit{method='trf'} \cite{Virtanen2020}. This bound-constrained trust-region algorithm was initialized at the default parameter values (`x0`) and restricted to the $[0,1]$ interval, with a maximum function evaluation budget (`max\_nfev`) set to match the imposed limit;

\end{itemize}

\subsection{Hierarchical Optimization}

Our main algorithm's components were implemented using \textit{scipy.optimize.minimize} \cite{Virtanen2020}.
\begin{itemize}
    \item Stiff subspace: It was performed using Powell's  \cite{powell1964};
    \item Sloppy subspace: It used Nelder-Mead methods \cite{nelder1965}, a simplex-based direct search.
\end{itemize}

Constraint Handling: The \textit{scipy} implementation of Powell and Nelder-Mead does not support the \textit{bounds} argument. Rather than using a wrapper, we enforced non-negative constraints directly within the simulator by taking the absolute value of the parameters before evaluation. While effective in our study, we note that more sophisticated solvers that explicitly accept \textit{bounds} could also be employed.

\section{Hierarchical Optimization Algorithm}
\label{app:hierarchical_algo}

Following the high-level description in Algorithm \ref{alg:hier_algo}, we provide a detailed specification for each of the functions referenced therein.

\begin{algorithm}
\caption{Hierarchical Optimization}
\begin{algorithmic}[1]
\Require Objective loss $\Phi(\cdot)$, initial guess $\theta^{(0)}$, max iters $N_{\max}$, stopping $\varepsilon_{\rm stop}$, Strategy $\in \{\text{Exact}, \text{Reduced}\}$

\State // \emph{Initialize Geometry}
\If{Strategy is \textbf{Exact}}
    \State Compute Hessian $H_{\rm GN}(\theta^{(0)})$ using Eq.\eqref{eq:Hessian_GN}
    \State $V^{(0)}_s, V^{(0)}_l \gets \text{eigendecompose} \left(H_{\rm GN}(\theta^{(0)}) \right)$ 
\Else \quad // \emph{Reduced Hessian Proxy}
    \State Generate random basis $\Omega \in \mathbb{R}^{n \times k}$
    \State Compute reduced Hessian $H_{\rm small}(\theta^{(0)}, \Omega)$ using Eq.\eqref{eq:reduced_hessian}
    \State $U_s, U_l \gets \text{eigendecompose}(H_{\rm small})$; \quad $V^{(0)}_s \gets \Omega U_s$; \quad $V^{(0)}_l \gets \Omega U_l$
\EndIf

\State $i \gets 1$, $\psi^{(0)} \gets 0$

\While{$i < N_{\rm max}$}

    \State // \emph{1) Stiff-subspace solve}
    \[
    \phi^{(i)} \gets \arg \min_{\phi} \Phi\left(\theta^{(i-1)} + V^{(i-1)}_s \phi + V^{(i-1)}_l \psi^{(i-1)} \right)
    \]
    \State $\theta^{(i')} \gets \theta^{(i-1)} + V^{(i-1)}_s \phi^{(i)}$

    \State // \emph{2) Sloppy Re-alignment (Optional)}
    \State Compute $Y_{l} \gets ({r(\theta^{(i')} + h V_l^{(i-1)}) - r(\theta^{(i')})})/{h}$ 
    \State $H_{l} \gets Y_{l}^T Y_{l}$; \quad $U_{l} \gets \text{eigendecompose}(H_{l})$
    \State $V'^{(i-1)}_l \gets V^{(i-1)}_l U_{l}$ 

    \State // \emph{3) Sloppy-subspace solve}
    \[
    \psi^{(i)} \gets \arg \min_{\psi} \Phi\left(\theta^{(i')} + V'^{(i-1)}_l \psi \right)
    \]

    \State // \emph{4) Update full parameter}
    \[
      \theta^{(i)} \;=\; \theta^{(i')} + V'^{(i-1)}_\ell\,\psi^{(i)}
    \]

    \State // \emph{5) Update Geometry}
    \If{Strategy is \textbf{Exact}}
        \State Compute $H_{\rm GN}(\theta^{(i)})$; update $V^{(i)}_s, V^{(i)}_l$
    \Else \quad // \emph{Reduced Proxy}
        \State Sample new $\Omega$; compute $H_{\rm small}(\theta^{(i)}, \Omega)$ 
        \State $U_s, U_l \gets \text{eigendecompose}(H_{\rm small})$
        \State Update $V^{(i)}_s \gets \Omega U_s$; \quad $V^{(i)}_l \gets \Omega U_l$
    \EndIf

    \State // \emph{6) Check stopping}
    \If{ $\mathrm{criteria}\left( V^{(i-1)}_s, V^{(i)}_s \right) < \varepsilon_{\rm stop}$ }
        \State \textbf{break}
    \EndIf
    \State $i \leftarrow i + 1$
\EndWhile

\State \Return $\theta^{(i)}$
\end{algorithmic}
\label{alg:hier_algo}
\end{algorithm}

\textbf{Gauss-Newton Hessian Computation}

For the computation of the \textit{Gauss-Newton} Hessian matrix $(H_{\rm GN})$ Eq.\eqref{eq:Hessian_GN}, we use the forward finite difference approximation. Moreover, to ensure the matrix is well-conditioned, we apply \textit{Tikhonov} regularization. Hence, we have:
\begin{equation}
    H_{\rm GN} \gets H_{\rm GN} + \lambda_{\rm reg} I
\end{equation}
where $I$ is the identity matrix.

\textbf{Stiff and Sloppy Subspace Decomposition}

We perform an eigendecomposition of the regularized $H_{\rm GN}$ to identify the stiff and sloppy subspaces of the parameter space. We denote  $\{ \lambda_i \}^n_{i=1}$ as the eigenvalues sorted in descending order, $(\lambda_1 \geq \lambda_2 \geq \dots \geq \lambda_n > 0)$, and, $\{ v_i\}^n_{i=1}$, their corresponding eigenvectors. 

The stiff subspace is defined as the space spanned by the $k_s$ eigenvectors, which collectively capture a fraction $\gamma$ of the total variance, in the style of Principal Component Analysis (PCA) \cite{Twarwat2016}. As presented in Table \ref{tab:hyperparameters}, we consider $\gamma = 0.90$.
$k_s$ is formally defined as the smallest integer satisfying:
\begin{equation}
    k_s = \min \left\{ k \in \{  1,\dots, n\} \bigg| \sum^k_{i=1} \lambda_i \geq \gamma \sum^n_{i=1} \lambda_i \right\}
\end{equation}
The remaining $n-k_s$ dimensions form the sloppy subspace.

\textbf{Reduced Sloppy Subspace for Optimization}

For our numerical optimization procedure, we further prune the sloppy subspace to improve sample efficiency. We only include sloppy directions whose eigenvalues are significant relative to the largest eigenvalue within the same subspace.

$\lambda_{k_s +1}$ is the largest sloppy eigenvalue, which we assume always belongs to the reduced sloppy subspace. Furthermore, we consider that the remaining sloppy directions $v_i$ are included in the reduced sloppy subspace only if their eigenvalue $\lambda_i$ meets a relative threshold $\tau$:
\begin{equation}
    \lambda_i \geq \tau \, . \,\lambda_{k_s + 1}~~~~\text{for} ~ i \in \{ k_s +2, \dots, n \}
    \label{eq:reduced_sloppy}
\end{equation}
The total number of sloppy directions retained is given by $k_l +1$, where $k_l$, is the number of eigenvalues satisfying Eq.\eqref{eq:reduced_sloppy}.

\textbf{Hyperparameters}

The optimization algorithm terminates when the convergence criteria (described in Section \ref{sec:Analysis}) $\varepsilon_{\rm stop}$ is met or after a maximum of $N_{\rm max}$ iterations. The specific hyperparameter values used for all results presented in this work are listed in Table 
\ref{tab:hyperparameters}.

\begin{table}[h!] 
    \centering 
        \begin{tabular}{c | l | c} Hyperparameter & Description & Value\\ 
        \hline 
        $N_{\rm max}$ & Number maximum of iterations & $50$ \\ 
        $\varepsilon_{\rm stop}$ & Convergence tolerance & $10^{-4}$\\ 
        $\lambda_{\rm reg}$	& Regularization strength & $10^{-6}$\\ 
        $\gamma$ & Stiff subspace variance threshold & $0.90$ \\ 
        $\tau$ & Reduced sloppy subspace threshold & $10^{-4}$ \end{tabular} \caption{Hyperparameter values used in Algorithm \ref{alg:hier_algo}.} \label{tab:hyperparameters} \end{table}

\section{Implicit Hessian Construction via Finite Differences}
\label{app:hierarchical_deriv}

We seek to compute the reduced Hessian matrix $H_{\rm small} \in \mathbb{R}^{k \times k}$ without explicitly forming the high-dimensional Hessian $H_{\rm GN} \in \mathbb{R}^{n \times n}$ or the Jacobian $J \in \mathbb{R}^{m \times n}$.
Recall that the reduced Hessian is the projection of the Gauss-Newton approximation onto the subspace spanned by the orthonormal basis $\Omega \in \mathbb{R}^{n \times k}$:
\begin{equation}
H_{\rm small} = \Omega^{T} H_{\rm GN} \Omega.
\end{equation}
Substituting the Gauss-Newton approximation $H_{\rm GN} = J^{T}J$, we obtain:
\begin{equation}
H_{\rm small} = \Omega^{T} (J^{T}J) \Omega = (J\Omega)^{T} (J\Omega).
\label{eq:factorized_small}
\end{equation}
Let $Y = J\Omega \in \mathbb{R}^{m \times k}$. The $i$-th column of $Y$, denoted $Y_i$, is the product of the Jacobian matrix and the $i$-th basis vector $\Omega_i$. By definition, this matrix-vector product $J\Omega_i$ represents the directional derivative of the residual vector $r(\theta)$ along the direction $\Omega_i$.

We can approximate this directional derivative efficiently using a forward finite difference scheme \cite{Pearlmutter1994}. For a small step size $h$, the Taylor expansion of the residual vector is:
\begin{equation}
r(\theta + h\Omega_i) \approx r(\theta) + J (h\Omega_i) = r(\theta) + h Y_i.
\end{equation}
Rearranging for $Y_i$, we get the computable approximation:
\begin{equation}
Y_i \approx \frac{r(\theta + h\Omega_i) - r(\theta)}{h}.
\end{equation}
Thus, we construct the matrix $Y$ column-by-column by performing $k$ additional simulator evaluations (one for each basis vector in $\Omega$). The reduced Hessian is then obtained efficiently via the matrix multiplication $H_{\rm small} = Y^{T}Y$, avoiding the $O(n)$ cost of the full Jacobian.

\section{Probabilistic Justification for Random Subspaces}
\label{app:random_projection_proof}

The validity of using a random subspace to find the stiffest direction relies on the sloppy spectral gap \cite{Transtrum_2011}. We assume the true Hessian has a dominant eigenvalue $\lambda_1$ (stiff) with eigenvector $v_1$, and a set of smaller eigenvalues bounded by $\lambda_{\rm sloppy}$.

We aim to maximize the Rayleigh Quotient $R_H(v) = (v^T H v) / (v^T v)$ within our random subspace $S$. Let $\tilde{v} \in S$ be a unit vector in our subspace. We can decompose $\tilde{v}$ into a component aligned with the true stiff direction $v_1$ and a component orthogonal to it (in the sloppy space):
\begin{equation}
\tilde{v} = \cos(\theta)v_1 + \sin(\theta)v_{\rm sloppy},
\end{equation}
where $\theta$ is the angle between the random vector $\tilde{v}$ and the true stiff direction $v_1$.
The curvature energy captured by this vector is:
\begin{equation}
R_H(\tilde{v}) = \tilde{v}^{T} H \tilde{v} \approx \lambda_1 \cos^2(\theta) + \lambda_{\rm sloppy} \sin^2(\theta).
\end{equation}
To successfully identify the stiff direction, the energy projected from the stiff mode must dominate the background energy from the sloppy modes. That is, we require the \textit{signal} ($\lambda_1 \cos^2 \theta$) to be greater than the \textit{noise} floor ($\lambda_{\rm sloppy}$):
\begin{equation}
\lambda_1 \cos^2(\theta) > \lambda_{\rm sloppy} \iff \frac{\lambda_1}{\lambda_{\rm sloppy}} > \frac{1}{\cos^2(\theta)}.
\end{equation}
For a random subspace of rank $k$ in a parameter space of dimension $n$, the expected squared overlap between a random vector and a fixed target vector is $\mathbb{E}[\cos^2(\theta)] = k/n$ \cite{halko2010}. Substituting this expectation yields the critical condition for success:
\begin{equation}
\frac{\lambda_1}{\lambda_{\rm sloppy}} > \frac{n}{k}.
\end{equation}
This inequality states that as long as the {stiffness ratio} ($\lambda_1 / \lambda_{\rm sloppy}$) exceeds the dimensionality ratio ($n/k$), the local energy landscape within the random subspace will be dominated by the projection of the true stiff direction \cite{Transtrum_2011}. This ensures that the eigenvector corresponding to the largest eigenvalue of $H_{\rm small}$ is a reliable proxy for $v_1$.

\section{Derivation of the Upper Bound for $\|\nabla \Phi(\theta_{\rm final})\|_2$}
\label{subsec:Upper_bound_Gradient}

We start by assuming that the algorithm has successfully converged after $N$ iterations. Hence, the stiff subspace stabilization criterion,
$(V^{(N-1)}_s \approx V^{(N)}_s)$ is fullfiled and we may conclude that  $P^{(N)}_{\rm s} \nabla \Phi (\theta_{\rm final}) = 0$, with $P_{s} = V_s \,V_s^{T}$ and $P_{l} = V_l \,V_l^{T}$.

Moreover, since the projection operators form a complete basis $(P_s + P_l = I)$, the final gradient must lie entirely in the sloppy subspace: $\nabla \Phi(\theta_{\rm final}) =  P_l \nabla \Phi(\theta_{\rm final}) $.
We can write $\nabla \Phi(\theta_{\rm final})$ in terms of the eigenbasis of $H_{\rm final}$ as
\begin{equation}
    \nabla \Phi(\theta_{\rm final}) = H_{\rm final} \, \Delta \tilde{\theta} ,
    \label{eq:express_grad_final}
\end{equation}
which is valid as $H_{\rm final}$ is a full-rank matrix, and $\Delta \tilde{\theta}$ corresponds to the needed displacement vector.
The Hessian $H_{\rm final}$ in its eigenbasis is given by $H_{\rm final} = \sum_i \lambda_i v_i v_i^T$. 
As a result, we can directly bound the norm of this vector as:
\begin{equation}
    \| \nabla \Phi(\theta_{\rm final}) \|_2 = \|P_l H_{\rm final}  \Delta \tilde{\theta} \|_2 \leq \|P_l H_{\rm final}\|_2 \cdot   \| \Delta \tilde{\theta}  \|_2.
    \label{eq:inequality_upper_bound}
\end{equation}
The Hessian $H_{\rm final}$ in its eigenbasis is given by $H_{\rm final} = \sum_i \lambda_i v_i v_i^T$, implying that
\begin{equation}
    P_l H_{\rm final} = \sum_{j \in \rm sloppy} v_j v_j^T 
    \sum_i \lambda_i v_i v_i^T = \sum_{j \in \rm sloppy} \sum_i \lambda_i v_j (v_j^T v_i)  v_i^T.
\end{equation}
Due to the orthonormality of the eigenvectors $(v_j^T v_i = \delta_{ij})$, the expression simplifies to:
\begin{equation}
    P_l H_{\rm final} = \sum_{j \in \rm sloppy} \lambda_j v_j v_j^T.
\end{equation}
As a result, the spectral norm of $P_l H_{\rm final} $ is given by:
\begin{equation}
    \| P_l H_{\rm final} \|_2 = \max_{j \in \rm sloppy} |\lambda_j | = |\lambda_{\rm max \, sloppy}|.
\end{equation}
Substituting back into inequality \eqref{eq:inequality_upper_bound}, we arrive at the following expression:
\begin{equation}
    \| \nabla \Phi(\theta_{\rm final}) \|_2 \leq  | \lambda_{\rm max \, sloppy} | \cdot \| \Delta \tilde{\theta} \|_2 
\end{equation}

The bound depends on the constructed $\Delta \tilde{\theta}$. To provide a physically meaningful result, we must relate this to the true displacement from the local minimum, $\Delta \theta = \theta_{\rm final} - \theta_{\min}$. The Taylor expansion around the local minimum provides the link:
\begin{equation}
    \nabla \Phi(\theta_{\rm final}) \approx H(\theta_{min}) \Delta \theta.
\end{equation}
By comparing this to Eq.\eqref{eq:express_grad_final}, we see that $\Delta \tilde{\theta} = (H_{\rm final})^{-1} H_{\rm min} \Delta \theta$. Assuming that $\theta_{\rm final}$ is close to $\theta_{\rm min}$, the Hessians are nearly identical $(H_{\rm final} \approx H_{\rm min})$, justifying the approximation $\Delta \tilde{\theta} \approx \Delta \theta$ .

\end{document}